\documentclass[11pt,letterpaper]{article}
\usepackage{emnlp2017}
\usepackage{times}
\usepackage{latexsym}

\emnlpfinalcopy
\newif\ifanonymous
\ifemnlpfinal
\anonymousfalse
\else
\anonymoustrue
\fi

\usepackage{amsmath}
\usepackage{amssymb}
\usepackage{graphicx}
\newcommand{\cev}[1]{\reflectbox{\ensuremath{\vec{\reflectbox{\ensuremath{#1}}}}}}
\usepackage{framed}
\usepackage{subfigure}
\usepackage{booktabs}
\usepackage{empheq}
\usepackage{array}
\usepackage{color}
\usepackage{enumitem}

\usepackage{multirow}

\newcommand{\ourModelName}{NQG}

\def\emnlppaperid{153}

\title{Neural Question Generation from Text: A Preliminary Study}

\author{Qingyu Zhou$^\dag$\thanks{\; Contribution during internship at Microsoft Research.} \hspace{0.15cm} Nan Yang$^\ddag$ \hspace{0.15cm} Furu Wei$^\ddag$ \hspace{0.15cm} Chuanqi Tan$^\sharp $ \hspace{0.15cm} Hangbo Bao$^\dag$ \hspace{0.15cm} Ming Zhou$^\ddag$ \\
	$^\dag$Harbin Institute of Technology, Harbin, China \\
	$^\ddag$Microsoft Research, Beijing, China \\
	$^\sharp $Beihang University, Beijing, China\\
	{\tt qyzhgm@gmail.com} \hspace{0.15cm} {\tt \{nanya, fuwei, mingzhou\}@microsoft.com}\\
	{\tt tanchuanqi@nlsde.buaa.edu.cn} \hspace{0.15cm} {\tt baohangbo@hit.edu.cn}
}

\date{}

\begin{document}

\maketitle

\begin{abstract}
	Automatic question generation aims to generate questions from a text passage where the generated questions can be answered by certain sub-spans of the given passage.
	Traditional methods mainly use rigid heuristic rules to transform a sentence into related questions.
	In this work, we propose to apply the neural encoder-decoder model to generate meaningful and diverse questions from natural language sentences.
	The encoder reads the input text and the answer position, to produce an answer-aware input representation, which is fed to the decoder to generate an answer focused question.
	We conduct a preliminary study on neural question generation from text with the SQuAD dataset, and the experiment results show that our method can produce fluent and diverse questions.
\end{abstract}

\section{Introduction}

Automatic question generation  from natural language text aims to generate questions taking text as input, which has the potential value of education purpose \citep{heilman2011automatic}.
As the reverse task of question answering, question generation also has the potential for providing a large scale corpus of question-answer pairs.

Previous works for question generation mainly use rigid heuristic rules to transform a sentence into related questions \citep{heilman2011automatic, chali2015towards}.
However, these methods heavily rely on human-designed transformation and generation rules, which cannot be easily adopted to other domains.
Instead of generating questions from texts, \citet{serban-EtAl:2016:P16-1} proposed a neural network method to generate factoid questions from structured data.

In this work we conduct a preliminary study on question generation from text with neural networks, which is denoted as the Neural Question Generation  (NQG) framework, to generate natural language questions from text without pre-defined rules.
The Neural Question Generation framework extends the sequence-to-sequence models by enriching the encoder with answer and lexical  features to generate answer focused questions.
Concretely, the encoder reads not only the input sentence, but also the answer position indicator and lexical features.
The answer position feature denotes the answer span in the input sentence, which is essential to generate answer relevant questions.
The lexical features include part-of-speech (POS) and named entity (NER) tags to help produce better sentence encoding.
Lastly, the decoder with attention mechanism \cite{bahdanau2014neural} generates an answer specific question of the sentence.

Large-scale manually annotated passage and question pairs play a crucial role in developing question generation systems.
We propose to adapt the recently released Stanford Question Answering Dataset (SQuAD) \citep{rajpurkar2016squad} as the training and development datasets for the question generation task.
In SQuAD, the answers are labeled as subsequences in the given sentences by crowed sourcing, and it contains more than 100K questions which makes it feasible to train our neural network models.
We conduct the experiments on SQuAD, and the experiment results show the neural network models can produce fluent and diverse questions from text.

\section{Approach}
\begin{figure}[ht]
	\centering
	\includegraphics[width=0.45\textwidth]{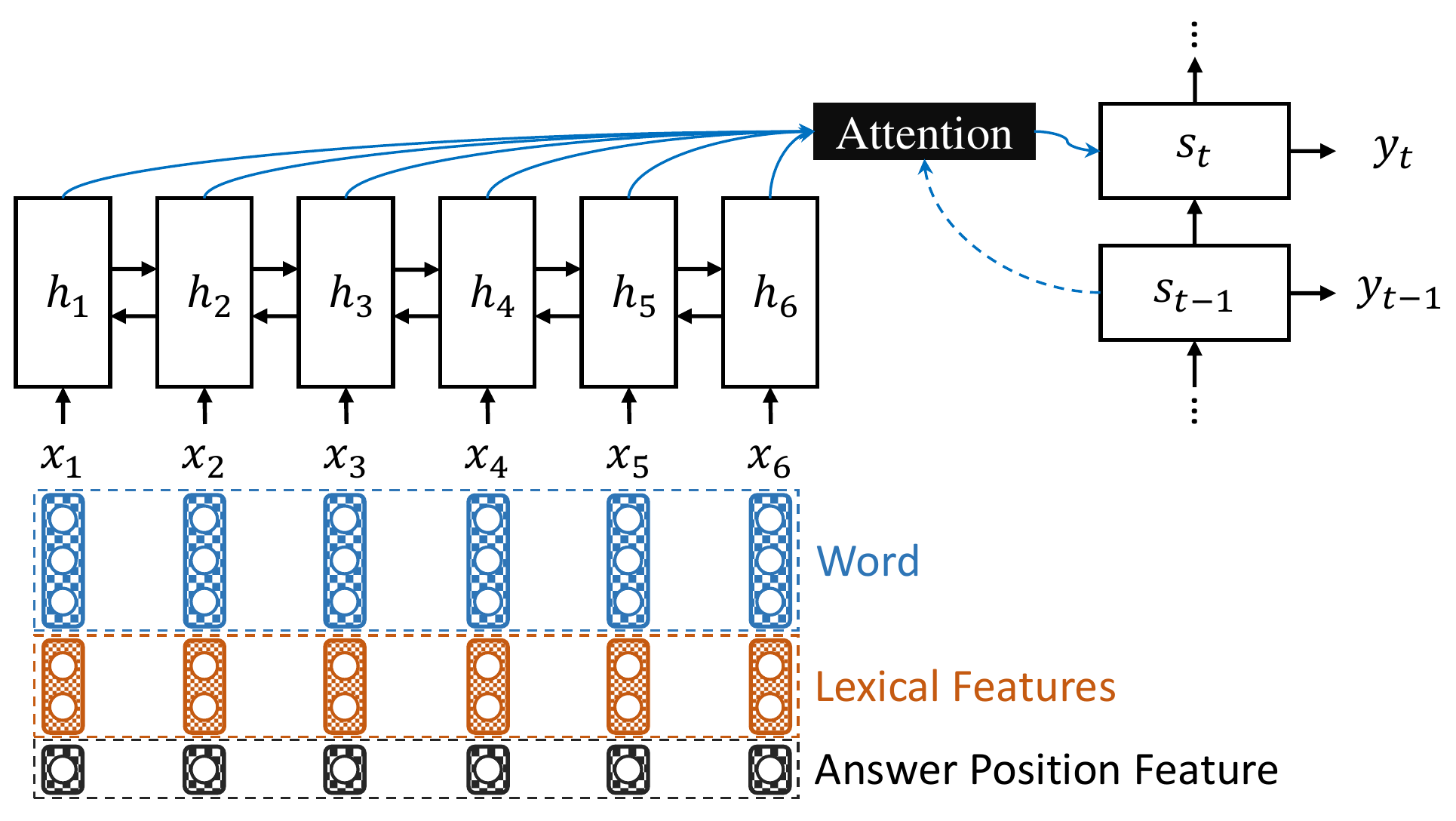}
	\caption{\label{fig:model}Overview of the Neural Question Generation (\ourModelName{}) framework.
	}
\end{figure}

In this section, we introduce the \ourModelName{} framework, which consists of a feature-rich encoder and an attention-based decoder.
Figure \ref{fig:model} provides an overview of our \ourModelName{} framework.

\subsection{Feature-Rich Encoder}
In the \ourModelName{} framework, we use Gated Recurrent Unit (GRU) \cite{cho-EtAl:2014:EMNLP2014} to build the encoder.
To capture more context information, we use bidirectional GRU (BiGRU) to read the inputs in both forward and backward orders.
Inspired by \citet{chen-manning:2014:EMNLP2014,nallapatiabstractive}, the BiGRU encoder not only reads the sentence words, but also handcrafted features, to produce a sequence of word-and-feature vectors.
We concatenate the word vector, lexical feature embedding vectors and answer position indicator embedding vector as the input of BiGRU encoder.
Concretely, the BiGRU encoder reads the concatenated sentence word vector, lexical features, and answer position feature, $ x = (x_{1}, x_{2}, \dots, x_{n}) $, to produce two sequences of hidden vectors, i.e., the forward sequence $ (\vec{h}_{1}, \vec{h}_{2}, \dots, \vec{h}_{n})  $ and the backward sequence $ (\cev{h}_{1}, \cev{h}_{2}, \dots, \cev{h}_{n}) $.
Lastly, the output sequence of the encoder is the concatenation of the two sequences, i.e., $ h_{i} = [\vec{h}_{i} ; \cev{h}_{i}] $.

\paragraph{Answer Position Feature}
To generate a question with respect to a specific answer in a sentence, we propose using answer position feature to locate the target answer.
In this work, the BIO tagging scheme is used to label the position of a target answer.
In this scheme, tag B denotes the start of an answer, tag I continues the answer and tag O marks words that do not form part of an answer.
The BIO tags of answer position are embedded to real-valued vectors throu and fed to the feature-rich encoder.
With the  BIO tagging feature, the answer position is encoded to the hidden vectors and used to generate answer focused questions.

\paragraph{Lexical Features}
Besides the sentence words, we also feed other lexical features to the encoder.
To encode more linguistic information, we select word case, POS and NER tags as the lexical features.
As an intermediate layer of full parsing, POS tag feature is important in many NLP tasks, such as information extraction and dependency parsing \citep{manning1999foundations}.
Considering that SQuAD is constructed using Wikipedia articles, which contain lots of named entities, we add NER feature to help detecting them.

\subsection{Attention-Based Decoder}
We employ an attention-based GRU decoder to decode the sentence and answer information to generate questions.
At decoding time step $ t $, the GRU decoder reads the previous word embedding $ w_{t-1} $ and context vector $ c_{t -1} $  to compute the new hidden state $ s_{t} $.
We use a linear layer with the last backward encoder hidden state $ \cev{h}_{1} $ to initialize the decoder GRU hidden state.
The context vector $ c_{t} $ for current time step $ t $ is computed through the concatenate attention mechanism \citep{luong-pham-manning:2015:EMNLP}, which matches the current decoder state $ s_{t} $ with each encoder hidden state $ h_{i} $ to get an importance score. The importance scores are then normalized to get the current context vector by weighted sum:
\begin{empheq}{align}
s_{t} &= \text{GRU}(w_{t-1}, c_{t - 1}, s_{t-1})\\
s_{0} &= \tanh (\mathbf{W}_{d}\cev{h}_{1} + b) \\
e_{t,i} &= v_{a}^{\top}\tanh(\mathbf{W}_{a}s_{t-1} + \mathbf{U}_{a}h_{i})\\
\label{eq:attentionProb}\alpha_{t,i} &= \frac{\exp (e_{t,i})}{\sum_{i=1}^{n}\exp (e_{t,i})}\\
c_{t} &= \sum_{i = 1}^{n} \alpha_{t,i}h_{i}
\end{empheq}

We then combine the previous word embedding $ w_{t-1} $, the current context vector $ c_{t} $, and the decoder state $ s_{t} $ to get the readout state $ r_{t} $.
The readout state is passed through a maxout hidden layer \citep{goodfellow2013maxout} to predict the next word with a softmax layer over the decoder vocabulary:
\begin{empheq}{align}
r_{t} &= \mathbf{W}_{r}w_{t-1} + \mathbf{U}_{r}c_{t} + \mathbf{V}_{r}s_{t}\\
m_{t} &= [\max\{r_{t, 2j-1}, r_{t, 2j}\}]^{\top}_{j = 1,\dots, d}\\
p(y_{t} &\vert y_{1}, \dots, y_{t-1}) = \text{softmax}(\mathbf{W}_{o}m_{t})
\end{empheq}
where $ r_{t} $ is a $ 2d $-dimensional vector.

\subsection{Copy Mechanism}
To deal with the rare and unknown words problem, \citet{gulcehre-EtAl:2016:P16-1} propose using pointing mechanism to copy rare words from source sentence.
We apply this pointing method in our \ourModelName{} system.
When decoding word $ t $, the copy switch takes current decoder state $ s_{t} $ and context vector $ c_{t} $ as input and generates the probability $ p $ of copying a word from source sentence:
\begin{empheq}{align}
p = \sigma(\mathbf{W}s_{t} + \mathbf{U}c_{t} +b)
\end{empheq}
where $ \sigma $ is sigmoid function. We reuse the attention probability in equation \ref{eq:attentionProb} to decide which word to copy.

\section{Experiments and Results}

%\subsection{Dataset}
We use the SQuAD dataset as our training data.
SQuAD is composed of more than 100K questions posed by crowd workers on 536 Wikipedia articles.
We extract sentence-answer-question triples to build the training, development and test sets\footnote{\label{label:releaseData}We re-distribute the processed data split and PCFG-Trans baseline code at {\ifanonymous an anonymous url for blind review.\else \url{http://res.qyzhou.me} \fi} }.
Since the test set is not publicly available, we randomly halve the development set to construct the new development and test sets.
The extracted training, development and test sets contain 86,635, 8,965 and 8,964 triples respectively.
We introduce the implementation details in the appendix.

%\subsection{Configuration}

We conduct several experiments and ablation tests as follows:
\begin{description}[noitemsep]
	\item[PCFG-Trans] The rule-based system\textsuperscript{\ref{label:releaseData}} modified on the code released by \citet{heilman2011automatic}.
	We modified the code so that it can generate question based on a given word span.
	\item[s2s+att] We implement a seq2seq with attention as the baseline method.
	\item[\ourModelName{}] We extend the s2s+att with our feature-rich encoder to build the \ourModelName{} system.
	\item[\ourModelName{}+] Based on \ourModelName{}, we incorporate copy mechanism to deal with rare words problem.
	\item[\ourModelName{}+Pretrain] Based on \ourModelName{}+, we initialize the word embedding matrix with pre-trained GloVe \citep{pennington2014glove} vectors.
%	\footnote{\url{http://nlp.stanford.edu/projects/glove/}}
	\item[\ourModelName{}+STshare] Based on \ourModelName{}+, we make the encoder and decoder share the same embedding matrix.
	\item[\ourModelName{}++] Based on \ourModelName{}+, we use both pre-train word embedding and STshare methods, to further improve the performance.
	\item[\ourModelName{}$ - $Answer] Ablation test, the answer position indicator is removed from \ourModelName{} model.
	\item[\ourModelName{}$ - $POS] Ablation test, the POS tag feature is removed from \ourModelName{} model.
	\item[\ourModelName{}$ - $NER] Ablation test, the NER feature is removed from \ourModelName{} model.
	\item[\ourModelName{}$ - $Case] Ablation test, the word case feature is removed from \ourModelName{} model.
\end{description}

\subsection{Results and Analysis}

We report BLEU-4 score \cite{papineni2002bleu} as the evaluation metric of our \ourModelName{} system.

\begin{table}[htbp]
		\small
	\begin{center}
		\begin{tabular}{lcc}
			\toprule
			\textbf{Model} &  Dev set & Test set  \\
			\midrule
			PCFG-Trans & 9.28 & 9.31 \\
			s2s+att & 3.01 & 3.06  \\
			\hline
			\hline
			\ourModelName{} & 10.06  & 10.13  \\
			\ourModelName{}+ & 12.30  & 12.18 \\
			\ourModelName{}+Pretrain & 12.80  & 12.69  \\
			\ourModelName{}+STshare & 12.92  &  12.80 \\
			\ourModelName{}++ & \textbf{13.27} & \textbf{13.29} \\
			\hline
			\hline
			\ourModelName{}$ - $Answer &  2.79 &  2.98  \\
			\ourModelName{}$ - $POS & 9.83  & 9.87 \\
			\ourModelName{}$ - $NER & 9.50 & 9.29 \\
			\ourModelName{}$ - $Case & 9.91  & 9.89 \\
			\bottomrule
		\end{tabular}
	\end{center}
	\caption{\label{tbl:bleu} BLEU evaluation scores of baseline methods, different \ourModelName{} framework configurations and some ablation tests.}
\end{table}

Table \ref{tbl:bleu} shows the BLEU-4 scores of different settings.
We report the beam search results on both development and test sets.
Our \ourModelName{} framework outperforms the PCFG-Trans and s2s+att baselines by a large margin.
This shows that the lexical features and answer position indicator can benefit the question generation.
With the help of copy mechanism, \ourModelName{}+ has a 2.05 BLEU improvement since it solves the rare words problem.
The extended version, \ourModelName{}++, has 1.11 BLEU score gain over \ourModelName{}+, which shows that initializing with pre-trained word vectors and sharing them between encoder and decoder help learn better word representation.

\paragraph{Human Evaluation}
We  evaluate the PCFG-Trans baseline and \ourModelName{}++ with human judges.
The rating scheme is, Good (3) - The question is meaningful and matches the sentence and answer very well;
Borderline (2) - The question matches the sentence and answer, more or less;
Bad (1) - The question either does not make sense or matches the sentence and answer.
We provide more detailed rating examples in the supplementary material.
Three human raters labeled 200 questions sampled from the test set to judge if the generated question matches the given sentence and answer span.
The inter-rater aggreement is measured with Fleiss' kappa \cite{fleiss1971measuring}.

\begin{table}[htbp]
	\small
	\begin{center}
		\begin{tabular}{lcc}
			\toprule
			\textbf{Model} &  AvgScore & Fleiss' kappa  \\
			\midrule
			PCFG-Trans & 1.42 & 0.50 \\
			\ourModelName{}++ & 2.18 & 0.46  \\
			\bottomrule
		\end{tabular}
	\end{center}
	\caption{\label{tbl:humanEva} Human evaluation results.}
%	\caption{\label{tbl:humanEva} Human evaluation results of PCFG-Trans baseline and \ourModelName{}++.}
\end{table}

Table \ref{tbl:humanEva} reports the human judge results.
The kappa scores show a moderate agreement between the human raters.
Our \ourModelName{}++ outperforms the PCFG-Trans baseline by 0.76 score, which shows that the questions generated by \ourModelName{}++ are more related to the given sentence and answer span.

\paragraph{Ablation Test}
The answer position indicator, as expected, plays a crucial role in answer focused question generation as shown in the \ourModelName{}$ - $Answer ablation test.
Without it, the performance drops terribly since the decoder has no information about the answer subsequence.

Ablation tests, \ourModelName{}$ - $Case, \ourModelName{}$ - $POS and \ourModelName{}$ - $NER,  show that word case, POS and NER tag features contributes to question generation.

\paragraph{Case Study}
Table \ref{tbl:example} provides three examples generated by \ourModelName{}++.
The words with underline are the target answers.
These three examples are with different question types, namely WHEN, WHAT and WHO respectively.
It can be observed that the decoder can `copy' spans from input sentences to generate the questions.
Besides the underlined words , other meaningful spans can also be used as answer to generate correct answer focused questions.
%With the answer position feature, \ourModelName{} can .

\begin{table}[htbp]
	\small
	\begin{center}
		\begin{tabular}{|lp{0.4\textwidth}|}
			\hline
			I: & in \underline{1226} , immediately after returning from the west , genghis khan began a retaliatory attack on the tanguts .\\
			G: & in which year did genghis khan strike against the tanguts ?\\
			O: & in what year did genghis khan begin a retaliatory attack on the tanguts ?\\
			\hline
			I: & in week 10 , manning suffered \underline{a partial tear of the} \underline{plantar fasciitis} in his left foot .\\
			G: & in the 10th week of the 2015 season , what injury was peyton manning dealing with ?\\
			O: & what did manning suffer in his left foot ?\\
			\hline
			I: & like the lombardi trophy , the `` 50 '' will be designed by \underline{tiffany \& co.} .\\
			G: & who designed the vince lombardi trophy ?\\
			O: & who designed the lombardi trophy ?\\
			\hline
		\end{tabular}
	\end{center}
	\caption{\label{tbl:example}Examples of generated questions, I is the input sentence, G is the gold question and O is the \ourModelName{}++ generated question. The underlined words are the target answers.}
\end{table}

\begin{figure}[ht]
	\centering
	\includegraphics[width=0.45\textwidth]{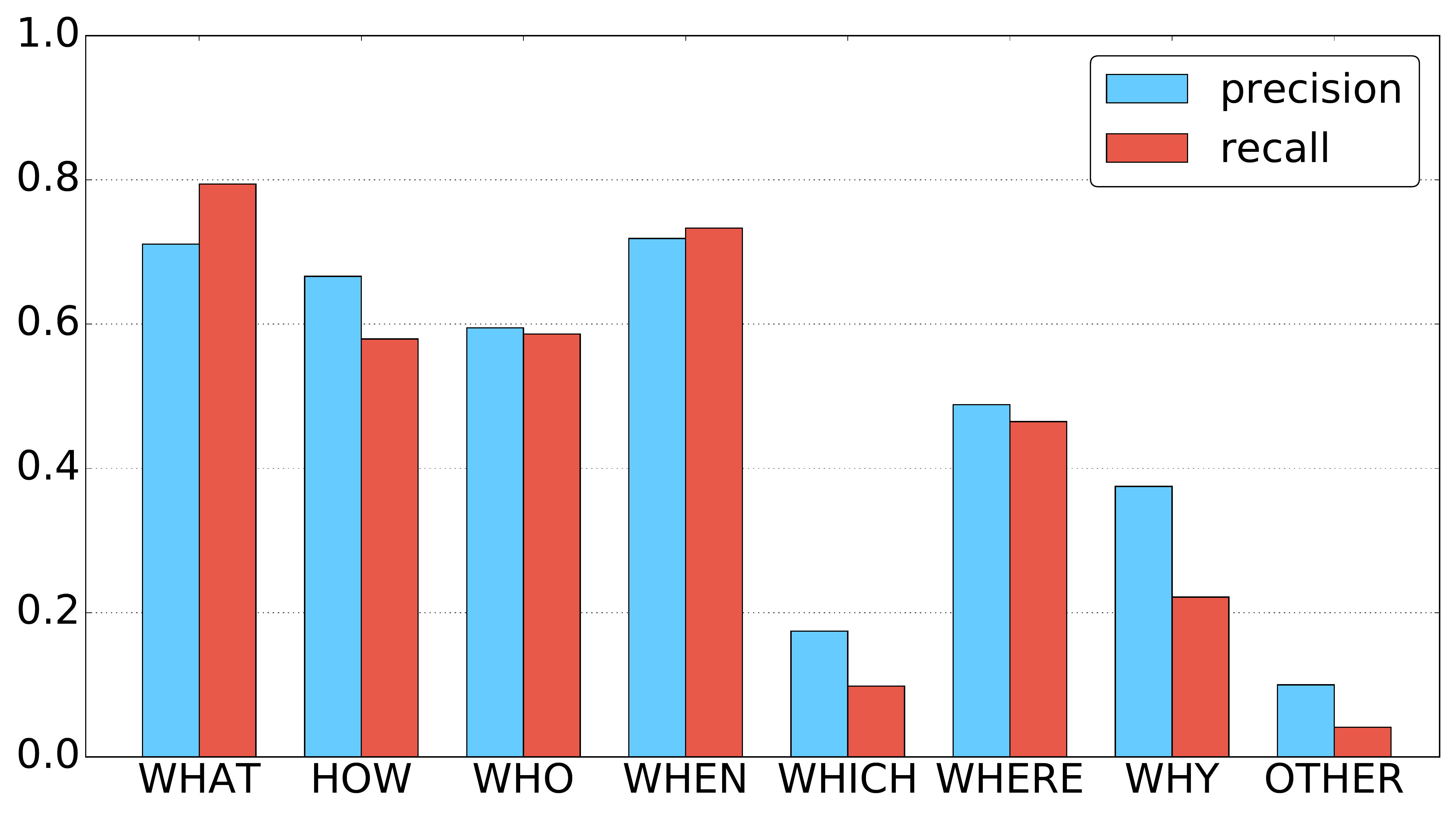}
	\caption{\label{fig:type}Precision and recall of question types.
	}
\end{figure}

\paragraph{Type of Generated Questions}
Following \citet{wang2016machine}, we classify the questions into different types, i.e., WHAT, HOW, WHO, WHEN, WHICH, WHERE, WHY and OTHER.\footnote{We treat questions `what country', `what place' and so on as WHERE type questions. Similarly, questions containing `what time', `what year' and so forth are counted as WHEN type questions.}
We evaluate the precision and recall of each question types.
Figure \ref{fig:type} provides the precision and recall metrics of different question types.
The precision and recall of a question type $ T $ are defined as:
\begin{empheq}{align}
\small\text{precision(T)} &= \small\frac{\text{\#(true T-type questions)}}{\text{\#(generated T-type questions)}}\\
\small\text{recall(T)} &= \small\frac{\text{\#(true T-type questions)}}{\text{\#(all gold T-type questions)}}
\end{empheq}
%\begin{center}
%	\small
%	\begin{empheq}{align}
%	\text{precision}(\small T) &= \small\frac{\#(\text{true } T \text{-type questions})}{\#(\text{generated } T \text{-type questions})}\\
%	\text{recall}(\small T) &= \small\frac{\#(\text{true } T \text{-type questions})}{\#(\text{all gold } T \text{-type questions})}
%	\end{empheq}
%\end{center}

For the majority question types, WHAT, HOW, WHO and WHEN types, our \ourModelName{}++ model performs well for both precision and recall.
For type WHICH, it can be observed that neither precision nor recall are acceptable.
Two reasons may cause this: a) some WHICH-type questions can be asked in other manners, e.g., `which team' can be replaced with `who'; b) WHICH-type questions account for about 7.2\% in training data, which may not be sufficient to learn to generate this type of questions. The same reason can also affect the precision and recall of WHY-type questions.

%\begin{table}[tbp]
%		\small
%	\begin{center}
%		\begin{tabular}{lrrrr}
%			\toprule
%			Type & Train & Test & \ourModelName{} & \ourModelName{}++ \\
%			\midrule
%			WHAT & 50.9 & 51.0 & 59.1 & 59.4 \\
%			HOW & 11.0 & 11.7 & 10.7 & 11.0 \\
%			WHO & 11.3 & 13.0 & 10.8 & 11.5 \\
%			WHEN & 11.2 & 9.8 & 9.7 & 9.6 \\
%			WHICH & 7.2 & 6.9 & 3.8 & 2.2 \\
%			WHERE & 5.4 & 5.2 & 4.7 & 5.0 \\
%			WHY & 1.4 & 1.6 & 1.0 & 1.0 \\
%			OTHER & 1.7 & 0.9 & 0.3 &0.3 \\
%			\bottomrule
%		\end{tabular}
%	\end{center}
%	\caption{\label{tbl:type}Percentage of question types in training, test, \ourModelName{} and \ourModelName{}++ generated questions.}
%\end{table}

\section{Conclusion and Future Work}
In this paper we conduct a preliminary study of natural language question generation with neural network models.
We propose to apply neural encoder-decoder model to generate answer focused questions based on natural language sentences.
The proposed approach uses a feature-rich encoder to encode answer position, POS and NER tag information.
Experiments show the effectiveness of our \ourModelName{} method.
In future work, we would like to investigate whether the automatically generated questions can help to improve question answering systems.

\bibliography{emnlp2017}
\bibliographystyle{emnlp_natbib}
%\clearpage
\appendix

%\newpage

\section{Implementation Details}

\subsection{Model Parameters}
We use the same vocabulary for both encoder and decoder.
The vocabulary is collected from the training data and we keep the top 20,000 frequent words.
We set the word embedding size to 300 and all GRU hidden state sizes to 512.
The lexical and answer position features are embedded to 32-dimensional vectors.
We use dropout \citep{srivastava2014dropout} with probability $ p = 0.5 $.
During testing, we use beam search with beam size 12.

\subsection{Lexical Feature Annotation}
We use Stanford CoreNLP v3.7.0 \citep{manning-EtAl:2014:P14-5} to annotate POS and NER tags in sentences with its default configuration and pre-trained models.

\subsection{Model Training}
We initialize model parameters randomly using a Gaussian distribution with Xavier scheme \citep{glorot2010understanding}.
We use a combination of Adam \citep{kingma2014adam} and simple SGD as our the optimizing algorithms.
The training is separated into two phases, the first phase is optimizing the loss function with Adam and the second is with simple SGD.
For the Adam optimizer, we set the learning rate $ \alpha = 0.001 $, two momentum parameters $ \beta_{1} = 0.9 $ and $ \beta_{2} = 0.999 $ respectively, and $ \epsilon=10^{-8} $.
We use Adam optimizer until the BLEU score on the development set drops for six consecutive tests (we test the BLEU score on the development set for every 1,000 batches).
Then we switch to a simple SGD optimizer with initial learning rate $ \alpha=0.5 $ and halve it if the BLEU score on the development set drops for twelve consecutive tests.
We also apply gradient clipping \citep{pascanu2013difficulty} with range $ [-5, 5] $ for both Adam and SGD phases.
To both speed up the training and converge quickly, we use mini-batch size 64 by grid search.

\section{Human Evaluation Examples}
We evaluate the PCFG-Trans baseline and \ourModelName{}++ with human judges.
The rating scheme is provided in Table \ref{tbl:humanEvaStand}.
\begin{table}[htbp]
	\begin{center}
		\begin{tabular}{lp{0.3\textwidth}}
			\toprule
			Score &  Rating scheme \\
			\midrule
			3: Good & The question is meaningful and matches the sentence and answer very well \\
			\hline
			2: Borderline & The question matches the sentence and answer, more or less \\
			\hline
			1: Bad & The question either does not make sense or matches the sentence and answer \\
			\bottomrule
		\end{tabular}
	\end{center}
	\caption{\label{tbl:humanEvaStand} Human rating scheme.}
\end{table}

The human judges are asked to label the generated questions if they match the given sentence and answer span according to the rating scheme and examples.
We provide some example questions with different scores in Table \ref{tbl:humanEvaExample}.
For the first score 3 example, the question makes sense and the target answer ``reason'' can be used to answer it given the input sentence.
For the second score 2 example, the question is inadequate for answering the sentence since the answer is about prime number.
However, given the sentence, a reasonable person will give the targeted answer of the question.
For the third score 1 example, the question is totally wrong given the sentence and answer.

\begin{table}[htbp]
	\begin{center}
		\centering
		\begin{tabular}{lp{0.38\textwidth}}
			\toprule
			Score &  Sentence and generated question \\
			\midrule
			\multirow{2}{*}{3} & \textbf{I:} -lsb- ... -rsb- for \underline{reason} is the greatest enemy that faith has ; it never comes to the aid of spiritual things . ''\\
		 	& 	\textbf{O:} what is the biggest enemy that faith has have ? \\
			\hline
			\multirow{2}{*}{2} & \textbf{I:}  in all other rows -lrb- a = 1 , 2 , 4 , 5 , 7 , and 8 -rrb- there are \underline{infinitely many prime numbers} . \\
			 & \textbf{O:} how many numbers are in all other rows ? \\
			\hline
			\multirow{2}{*}{1} & \textbf{I:}  while genghis khan never conquered all of china , his \underline{grandson} kublai khan completed that conquest and established the yuan dynasty that is often credited with re-uniting china .\\
			& \textbf{O:} who did kublai khan defeat that conquered all of china ?\\
			\bottomrule
		\end{tabular}
	\end{center}
	\caption{\label{tbl:humanEvaExample} Human rating scheme examples, \textbf{I} is input, \textbf{O} is output, the underlined words are target answers.}
\end{table}

\end{document}